\documentclass[sigconf]{aamas}

\usepackage{booktabs}

\usepackage{flushend}
\acmDOI{doi}  
\acmISBN{}  
\acmConference[Conference]{21}{October}{2019}  
\acmYear{2019}  
\copyrightyear{2019}  
\acmPrice{}  

\usepackage[utf8]{inputenc}
\usepackage{amsmath,amssymb}
\usepackage{algorithm}
\usepackage{algorithmicx}
\usepackage{algcompatible}
\usepackage{algpseudocode}
\usepackage{graphicx}
\usepackage{subcaption}
\usepackage{multirow}

\DeclareMathOperator*{\argmax}{arg\,max}

\begin{document}

\title{A New Framework for Multi-Agent Reinforcement Learning -- Centralized Training and Exploration with Decentralized Execution via Policy Distillation}

\author{Gang Chen \\ {\small Victoria University of Wellington, New Zealand}}  

\begin{abstract}
Deep reinforcement learning (DRL) is a booming area of artificial intelligence. Many practical applications of DRL naturally involve more than one collaborative learners, making it important to study DRL in a multi-agent context. Previous research showed that effective learning in complex multi-agent systems demands for highly coordinated environment exploration among all the participating agents. Many researchers attempted to cope with this challenge through learning centralized value functions. However, the common strategy for every agent to learn their local policies directly often fail to nurture strong inter-agent collaboration and can be sample inefficient whenever agents alter their communication channels. To address these issues, we propose a new framework known as centralized training and exploration with decentralized execution via policy distillation. Guided by this framework and the maximum-entropy learning technique, we will first train agents' policies with shared global component to foster coordinated and effective learning. Locally executable policies will be derived subsequently from the trained global policies via policy distillation. Experiments show that our new framework and algorithm can achieve significantly better performance and higher sample efficiency than a cutting-edge baseline on several multi-agent DRL benchmarks.
\end{abstract}

\keywords{Deep Reinforcement Learning; Multi-Agent Learning; Policy Distillation}  

\maketitle

\section{Introduction}
\label{sec-intro}

\emph{Deep reinforcement learning} (DRL) is a booming area of artificial intelligence \cite{mnih2015,mnih2016icml,haarnoja2018icml}. Advanced DRL technologies, mostly under single-agent settings, have achieved outstanding success in a wide range of fields such as renewable energy \cite{glavic2017}, self-driving vehicle \cite{sallab2017}, image processing \cite{caicedo2015}, robotics control \cite{gu2016icra}, and intelligent healthcare \cite{pineau2009}.

Many practical applications of DRL naturally involve more than one interdependent learners. For example, in multi-robot control tasks and multi-player computer games, multiple agents must learn to collaborate and jointly solve a challenging problem \cite{long2018,rashid2018}. There are several main reasons to study \emph{multi-agent DRL} (MADRL) \cite{hernandez2018survey,chen2005maco}. For example, partitioning the problem domain according to the role played by each agent can result in a similar effect as imposing hierarchical abstractions in both time and space, contributing positively to improved learning efficiency and effectiveness. MADRL also stands for a major paradigm for training agents to interact with each other productively \cite{sukhbaatar2017}. It is essential for the central theme of this paper that features a group of cooperative agents. Each agent has limited problem solving capabilities and must learn to work as a team. Previous research works have clearly showed that studying such MADRL problems possesses both theoretical and practical significance \cite{panait2005,shoham2007,shoham2003}.

Researchers have attempted to tackle MADRL problems by using single-agent DRL algorithms with success \cite{tampuu2017}. Despite of promising results reported recently, there are a few core issues that are inherent to MADRL and should be addressed carefully. For example, \emph{environment non-stationarity} hinders stable DRL because individual agents can no longer perceive their environment as being stationary since it is also influenced by other agents' activities. Efforts have been made to address this challenge by extending conventional experience replay mechanism with various new safety measures \cite{foerster2017,tesauro2004}. This idea has also promoted the wide adoption of the fundamental paradigm known as \emph{centralized training with decentralized execution} \cite{lowe2017nips,kraemer2016}. Moreover, interesting techniques, such as model-guided policy training and agent communication, have been proposed to handle the common situation where every agent only has partial view of their learning environment and other agents \cite{raileanu2018,lazaridou2017,foerster2016comm,sukhbaatar2016,peng2017wang}.

Our research is inspired by the understanding that effective learning in complex multi-agent systems demands for highly coordinated environment exploration among all the participating agents \cite{matignon2012}. In particular agents must carefully select their local actions based on a global view of the multi-agent system in order to achieve system-wide outcomes. However, the common strategy for agents to directly learn their \emph{local policies} cannot easily facilitate strong coordination since these policies only have restricted access to agents' local observations. Such lack of observability can be mitigated with the introduction of inter-agent communication channels that allow agents to share instant messages and expand their environment knowledge. However, this will inevitably lead to increased learning complexity, especially when agents are struggling to learn how to communicate effectively and how to maximize their \emph{expected long-term rewards} at the same time. Moreover, additional DRL cycles must be performed to collect new environment samples and re-train local policies whenever the communication channel is altered (e.g., when more information can be exchanged among agents per time step), further reducing sample efficiency.

In order to achieve coordinated environment exploration and high sample efficiency, we propose a new framework known as \emph{Centralized Training and Exploration with Decentralized execution via policy Distillation} (i.e., CTEDD) with strong emphasis on global information sharing. In other words, before building up their local policies, agents first train their \emph{global policies} to process full state input through a shared global \emph{deep neural network} (DNN). Such global DNN paves the way for coordinated action sampling and environment exploration. Specifically, with the help of the maximum-entropy RL technique \cite{haarnoja2018icml}, all agents will be trained to collectively decide when to explore aggressively and when to focus on exploiting policies learned so far, resulting in balanced trade-off between exploration and exploitation.

Global policy training as described above can improve both sample efficiency and learning performance. However, the learned policies are not amenable to decentralized execution due to their reliance on global information (rather than agents' local observations). Guided by CTEDD, we propose to adopt a key technique known as \emph{policy distillation} to derive locally executable policies for every agent from the global policies \cite{rusu2016}. Since policy distillation is a supervised learning task, it is more reliable and controllable than MADRL, ideal for training local policies with flexible inter-agent communication capabilities. Moreover, because local policies are trained only on environment samples collected \emph{a priori} by using global policies, no extra sampling cost will be incurred, even after altering the communication channels.

In the past, policy distillation has been utilized mainly for single-agent DRL \cite{rusu2016,berseth2019}. In MADRL, the technique was introduced to support multi-task learning and learning a single shared policy network for all agents based on multiple pre-trained networks \cite{omidshafiei2017}. In this paper we will pursue a different direction of applying policy distillation to training agents' local policies with high sample efficiency.

To drive our study of the CTEDD framework, similar to \cite{lowe2017nips,iqbal2018}, we focus on MADRL problems with multi-dimensional continuous action spaces. Various policy gradient learning methods have been utilized for tackling such continuous action learning problems with success \cite{schulman2015icml,schulman20171,wu2017nips,haarnoja2017}. Among them, one DRL algorithm named \emph{Deep Deterministic Policy Gradient} (DDPG) has attracted substantial attention due to its leading performance and high sample efficiency on many difficult RL benchmarks \cite{lillicrap2015}. In particular, a recent multi-agent extension of DDPG, known as MADDPG \cite{lowe2017nips}, will serve as the baseline algorithm in this paper. Building on MADDPG, we will study the key advantages of CTEDD over the prevalent approach of learning agents' local policies directly. Empirical comparison with MADDPG confirms that CTEDD is more sample efficient and effective, while guaranteeing decentralized execution of the learned policies.

MADDPG is considered suitable for this research because it shares the common goal with CTEDD to explicitly train agents' policies rather than indirectly deriving these policies from previously trained value functions. MADDPG performs highly competitively to other state-of-the-art MADRL algorithms such as COMA that also rely on policy gradient training techniques \cite{foerster2018}. Moreover, MADDPG is capable of training local policies with communication capabilities, provided that gradients can be passed through communication channels during learning. While we focus on expanding MADDPG with CTEDD in this paper, CTEDD is general enough to be applied to other MADRL algorithms to improve policy training. CTEDD is also highly compatible with other recent enhancements to MADDPG, which will be explained in Section \ref{sec-rw}. 

\section{Related Works}
\label{sec-rw}

The most straightforward approach to MADRL via agents that learn independently may not always produce satisfactory results. As demonstrated in \cite{lowe2017nips}, policy gradient methods often exhibit high level of variance when agent coordination is essential to effective MADRL. Meanwhile, na\"ive application of Deep Q-Networks (DQNs) can lead to unpredictable and diverging outcomes. To address this issue, many interesting research works have been reported recently in the literature. We will present a short summary of closely related studies organized into three categories, i.e., coordinated environment exploration, policy gradient methods, and learning inter-agent communication.

Researchers are well versed of the importance for agents to explore their learning environment in a coordinated manner. In recent years, many interesting techniques, including parameter sharing \cite{sunehag2018}, the leniency mechanism \cite{palmer2018}, hysteretic Q-function update \cite{foerster2017}, and potential reward shaping methods \cite{devlin2014}, have been proposed to promote concerted and optimistic environment exploration among cooperating agents. Majority of these techniques were developed for MADRL in discrete action spaces and cannot be immediately applied to problems with continuous actions. Meanwhile, their effectiveness often depends on useful domain knowledge. For example, \cite{devlin2014} assumes that agents can always quantify the desirability of any given states towards the learning goal.

In an effort to nurture strong collaboration, numerous MADRL algorithms propose to adopt centralized value functions (aka. \emph{critics} under the actor-critic DRL framework). For example, COMA employs a central critic to train distributed actors \cite{foerster2018}. Decomposible DNNs were further introduced by Sunehag \emph{et al.} to model value functions with improved learning scalability \cite{sunehag2018}. As evidenced in these works, central learning can become too complex to handle with an increasing number of agents. To address this issue, Q-MIX introduced a network architecture with mixed global and local components to reduce the complexity of central training \cite{rashid2018}. We follow a similar idea in this paper. However, in addition to training central critics, we will simultaneously train central actors for multi-agent action sampling in continuous spaces. In comparison, Q-MIX only supports discrete actions.

Policy gradient methods for MADRL are attracting increasing attention in the research community. In \cite{lowe2017nips}, MADDPG was proposed to train agents' local policies in a mixed environment with both cooperative and competitive agents. A very recent work has further applied adversarial training techniques to improving the robustness of learned policies in worst case scenarios \cite{li2019}. Meanwhile, the multi-actor-attention-critic (MAAC) algorithm is developed in \cite{iqbal2018} to improve the scalability and effectiveness of critic learning in MADDPG with a new attention mechanism. Our CTEDD framework is compatible with (and orthogonal to) this research because (1) CTEDD focuses on enhancing agent collaboration during policy training; and (2) although critic training is not the main target, CTEDD can easily adopt the attention mechanism introduced by MAAC to train critics whenever necessary. In other words, MAAC can be extended with the new policy training approach developed in CTEDD. Hence instead of training local policies directly as in MAAC, we can follow CTEDD to train policy networks with mixed global and local structures inspired by Q-MIX. Local policies can be further derived from the trained global policies via policy distillation. However such combination of MAAC and CTEDD is beyond the scope of this paper.

A fundamental property of multi-agent interaction is marked by emergent communication. Learning to communicate is important when agents can only observe their environment partially. Representative algorithms like RIAL and DIAL rely on DQN to drive inter-agent communication through discrete signalling channels \cite{foerster2016comm}. Meanwhile, CommNet considers a continuous vector channel and permits multi-cycle communication per time step \cite{sukhbaatar2016}. A latest research introduces BicNet to facilitate latent information exchange via bidirectional recurrent neural networks (RNNs) \cite{peng2017wang}. Without relying on sophisticated network architectures, in this paper, we study the use of simple restrictive agent-to-agent communication channels that are more feasible in resource-limited applications. Moreover, different from many previous works, our algorithm postpones communication learning to the policy distillation stage. This helps to significantly reduce the difficulty involved in training agents' policies through MADRL algorithms such as MADDPG.

\section{Preliminaries}
\label{sec-bac}

\subsection{Problem Definition}
\label{sub-bac-pd}

In this paper, the \emph{cooperative} MADRL problem is defined as a natural extension of the conventional \emph{Markov Decision Process} (MDP) \cite{sutton1998book} in a partially observable multi-agent system. Specifically, a total of $N$ agents are jointly located in the same learning environment $\mathbb{E}$. The system-wide information about $\mathbb{E}$ is captured through a set of states $\mathbb{S}$. At any time step $t$, every agent can obtain a local observation of state $s_t\in\mathbb{S}$ via their individual observation functions $o_i(s_t)$, with $1\leq i\leq N$. Since $o_i(s_t)$ may lose vital information embedded in $s_t$, to cope with agents' restricted observability, we establish inter-agent communication channels that allow every agent to send a message to all other agents at each time step and use $m^t_i$ to denote the message generated by agent $A_i$ at time $t$. Meanwhile an agent can impose control over $\mathbb{E}$ and other agents by performing a series of actions over time. The local action $a^t_i$ to be performed by agent $A_i$ at any time $t$ must be selected from a bounded multi-dimensional action space $\mathbb{A}_i$.

To guide local action selection, every agent $A_i$ employs a deterministic \emph{local policy} $\hat{\pi}_{\theta'_i}$ parameterized by $\theta'_i$ that maps its local observation $o_i(s_t)$ of state $s_t$ and all messages received $\{m^t_{j\neq i}\}_{j=1}^N$ at time $t$ to a specific action $a^t_i\in\mathbb{A}_i$. Conditional on all the actions performed by each agent, i.e., $\{a^t_i\}_{i=1}^N$, the environment $\mathbb{E}$ transits to a new state $s_{t+1}$ according to the probability distribution $\mathcal{T}(s_{t+1}|s_t,\{a^t_i\}_{i=1}^N)$. Meanwhile, $\mathbb{E}$ produces a scalar reward $r(s_t,\{a^t_i\}_{i=1}^N)$ to all learning agents. Under this framework, the ultimate goal of MADRL is to identify a collection of local policies so as to jointly maximize the expected long-term rewards $R$, as described mathematically below.
\begin{equation}
\argmax_{\{\theta'_i\}_{i=1}^N} R\left(\{\hat{\pi}_{\theta'_i}\}_{i=1}^N \right)
\label{eq-pro-def}
\end{equation}
\noindent
with
\[
\begin{split}
& R\left(\{\hat{\pi}_{\theta'_i}\}_{i=1}^N\right)=\\
& \underset{s_t,\{a^t_i\}_{i=1}^N\sim \{\hat{\pi}_{\theta'_i}\}_{i=1}^N}{\mathrm{E}}
\left[ \sum_{t} \gamma^t r\left(s_t,\{\hat{\pi}_{\theta'_i}(o_i(s_t),\{m^t_{j\neq i}\}_{j=1}^N) \}_{i=1}^N \right) \right]
\end{split}
\]
\noindent
where $\gamma\in[0,1)$ is a discount factor. 

Directly learning local policies $\hat{\pi}_{\theta'_i}$ described above can be slow and sample inefficient. Following the idea of CTEDD, we introduce a \emph{global policy} $\tilde{\pi}_i$ for every agent $A_i$ as a combination of the deterministic part (i.e., $\pi_{\theta_i}$) and the stochastic part (i.e., $\sigma_{\omega_i}$), parameterized respectively by $\theta_i$ and $\omega_i$. Both parts are direct functions of state $s_t$, as shown below.
\[
\tilde{\pi}_i(s_t) =\pi_{\theta_i}(s_t)+\epsilon\cdot \sigma_{\omega_i}(s_t)
\]
\noindent
where $\epsilon$ is sampled from spherical Gaussian and $\sigma_{\omega_i}$ controls the scale of random local exploration by agent $A_i$. For simplicity, we also use $\tilde{\pi}_i(s_t,a^t_i)$ to denote the probability for agent $A_i$ to perform action $a^t_i$ in state $s_t$. Finding the optimal global policies $\{\tilde{\pi}_i\}_{i=1}^N$ hence becomes the new goal for the \emph{centralized training and exploration stage} of our proposed algorithm (see Subsection \ref{sub-met-cte}).

\subsection{Value Function Learning}
\label{sub-bac-vfl}

Learning value functions represented as DNNs has been popularly supported by many DRL and MADRL algorithms. In line with the MADRL problem introduced in Subsection \ref{sub-bac-pd}, the Q-function is defined as
\[
Q^{\tilde{\pi}}(s,\{a_i\}_{i=1}^N)=R\left(
\{\tilde{\pi}_i\}_{i=1}^N \left|
s_0=s, \{a^0_i\}_{i=1}^N=\{a_i\}_{i=1}^N
\right.
\right)
\]
Given an \emph{experience replay buffer}, $Q^{\tilde{\pi}}$ can be learned through minimizing the loss below over a batch of samples $\mathcal{B}$ collected from the buffer.
\begin{equation}
\begin{split}
& \mathcal{L}_Q= \\
& \sum_{\mathcal{B}}\left( Q^{\tilde{\pi}}(s_t,\{a^t_i\}_{i=1}^N)- r_t-\gamma Q^{\tilde{\pi}}(s_{t+1},\{\tilde{\pi}_i(s_{t+1})\}_{i=1}^N) \right)^2
\end{split}
\label{equ-loss-q}
\end{equation}
\noindent
Every sample assumes the form $<s_t,s_{t+1},\{a^t_i\}_{i=1}^N, r_t>$.

\subsection{Stochastic and Deterministic Policy Gradient Algorithms}
\label{sub-bac-alg}

Policy gradient algorithms can be utilized to directly train the global policy $\tilde{\pi}_i$ associated with each agent $A_i$. Since global policies are stochastic, according to the \emph{stochastic policy gradient theorem} \cite{sutton2000}, the policy gradient with respect to the stochastic part (i.e., $\sigma_{\omega_i}$ of $\tilde{\pi}_i$) of these policies can be determined according to
\begin{equation}
\begin{split}
&\nabla_{\omega_i}R\left(\{\tilde{\pi}_j\}_{j=1}^N \right)=\\
&\underset{s_t,\{a^t_j\}_{j=1}^N\sim \{\tilde{\pi}_j\}_{j=1}^N}{\mathrm{E}}\left[ \nabla_{\omega_i} \log \tilde{\pi}_i(s_t,\{a^t_j\}_{j=1}^N) Q^{\tilde{\pi}}(s_t, \{a^t_j\}_{j=1}^N)\right]
\end{split}.
\label{equ-pgt}
\end{equation}
As \eqref{equ-pgt} depends strongly on state samples, carefully controlled environment exploration is essential for effective learning. CTEDD adopts a global policy training method enhanced by maximum-entropy RL techniques to address this issue effectively (see Subsection \ref{sub-met-cte}). In comparison to other policy gradient learning methods such as PPO \cite{schulman20171} and TRPO \cite{schulman2015icml}, maximum-entropy RL is more suitable for this research because (1) it is designed to train stochastic policies for effective environment exploration; (2) it supports the reparameterization trick that underpins MADDPG; and (3) it learns reliably well on both on-policy and off-policy environment samples and has achieved leading performance on many continuous-action RL benchmarks \cite{haarnoja2018icml}.

In addition to \eqref{equ-pgt}, following MADDPG, the deterministic part $\pi_{\theta_i}$ of the global policy $\tilde{\pi}_i$ for every agent $A_i$ will be trained according to the \emph{deterministic policy gradient theorem} \cite{silver2014}, as shown below
\begin{equation}
\begin{split}
& \nabla_{\theta_i}R\left(\{\tilde{\pi}_j\}_{j=1}^N \right)\approx \\
& \frac{1}{\|\mathcal{B}\|}\sum_{\mathcal{B}}\left[ \nabla_{\theta_i}\pi_{\theta_i}(s_t)\nabla_{a_i}Q^{\tilde{\pi}}(s_t,\{a_j\}_{j=1}^N) |_{a_i=\pi_{\theta_i}(s_t)}\right]
\end{split}.
\label{equ-dpg}
\end{equation}

\begin{algorithm}[!ht]
 \begin{algorithmic}[1]
 \State {\bf Input}: DNNs with mixed global and local components that implement $\{\tilde{\pi}_{\theta_i}\}_{i=1}^N$, a centralized Q-network, and an experience replay buffer that stores past environment samples for centralized training
 \State {\bf for} each learning episode {\bf do}:
 \State \ \ \ \ Obtain initial state $s_0$ from environment
 \State \ \ \ \ {\bf for} $t=0,\ldots$ until end of episode {\bf do}:
 \State \ \ \ \ \ \ \ \ Sample local actions for each agent according to
 \Statex \ \ \ \ \ \ \ \ $\{\tilde{\pi}_{\theta_i}\}_{i=1}^N$
 \State \ \ \ \ \ \ \ \ Perform sampled local actions
 \State \ \ \ \ \ \ \ \ Add environment sample $<s_t,s_{t+1},\{a^t_i\}_{i=1}^N, r_t>$
 \Statex \ \ \ \ \ \ \ \ into the replay buffer
 \State \ \ \ \ \ \ \ \ {\bf if} learning interval is reached {\bf do}:
 \State \ \ \ \ \ \ \ \ \ \ \ \ Sample mini-batch $\mathcal{B}$
 \State \ \ \ \ \ \ \ \ \ \ \ \ Train centralized Q-network based on $\mathcal{B}$
 \State \ \ \ \ \ \ \ \ \ \ \ \ Train $\{\tilde{\pi}_{\theta_i}\}_{i=1}^N$ based on Q-network and $\mathcal{B}$
 \State {\bf for} each policy distillation iteration {\bf do}:
 \State \ \ \ \ Sample mini-batch $\mathcal{B}$ chronologically
 \State \ \ \ \ Prepare training data in the form
 \Statex \ \ \ \ $<s_t,\{\pi_{\theta_i}(s_t)\}_{i=1}^N>$ based on $\mathcal{B}$
 \State \ \ \ \ Train local policies $\{\hat{\pi}_{\theta_i'}\}_{i=1}^N$ to minimize the loss 
 \Statex \ \ \ \ in \eqref{equ-loss-d} with multiple repetitions
 \end{algorithmic}
\caption{An overview of the CTEDD algorithm.}
\label{alg-1}
\end{algorithm}

\section{Methods}
\label{sec-met}

We have the goal in this section to develop a new algorithm for MADRL. The same as DDPG, our algorithm aims to learn locally executable policies for each agent in a multi-agent system. However, instead of training such local policies directly, our algorithm realizes the idea of CTEDD by dividing policy training into two consecutive stages, i.e., the \emph{centralized training and exploration} stage and the \emph{policy distillation} stage. For simplicity, in the sequel, our algorithm will be called the CTEDD algorithm.

An overview of the CTEDD algorithm is presented in Algorithm \ref{alg-1}. As described in Algorithm \ref{alg-1}, CTEDD first initializes and trains global policy networks and a centralized Q-network. All networks receive full state input. The global policy networks also support local action selection by each agent through their individual local policy components. More details can be found in Subsection \ref{sub-met-cte}. After training the global policies, the deterministic parts of $\{\tilde{\pi}_i\}_{i=1}^N$, i.e., $\{\pi_{\theta_i}\}_{i=1}^N$, subsequently serve as the teacher policies to supervise the training of agents' local policies $\{\hat{\pi}_{\theta_i'}\}_{i=1}^N$ through policy distillation.

\subsection{Centralized Policy Training and Exploration}
\label{sub-met-cte}

\begin{figure}[ht!]
\centering
\includegraphics[width=0.95\columnwidth]{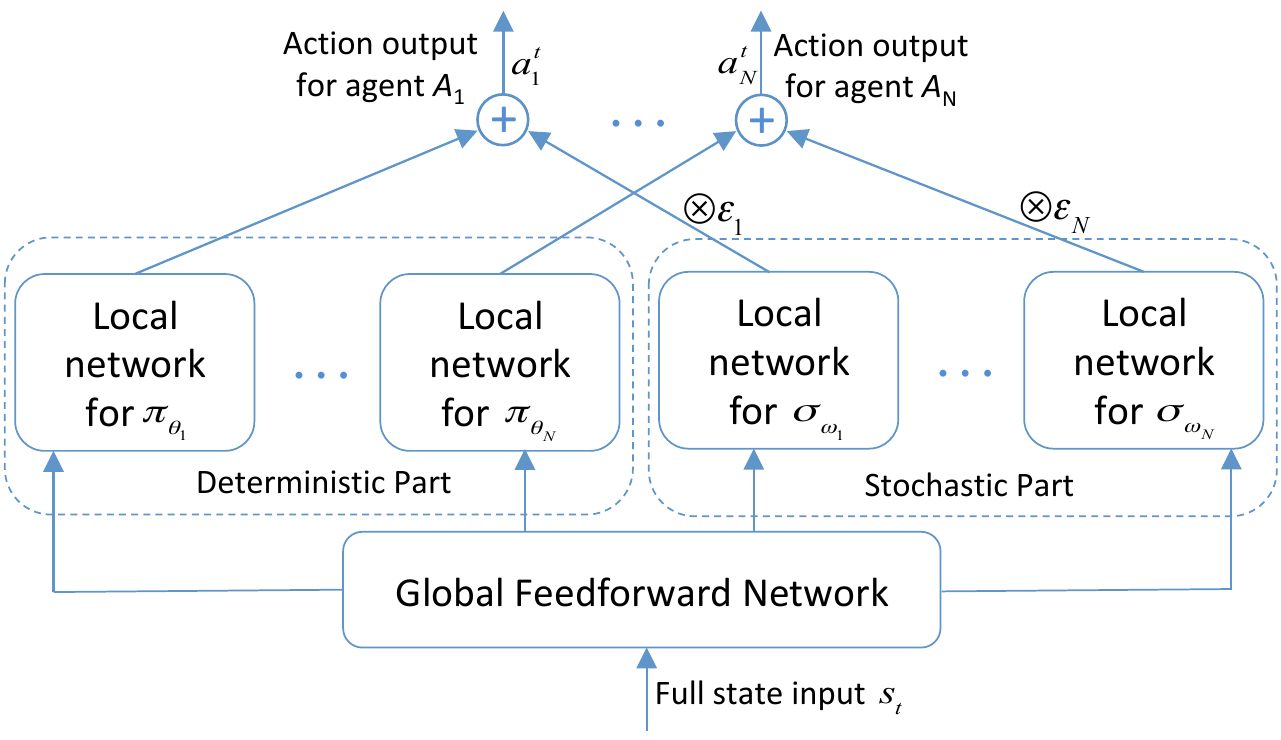}
\caption{The global policy network design with mixed global and local components.}
\label{fig-glo-pol}
\end{figure}

We design the neural network architecture for the global policies $\{\tilde{\pi}_i\}_{i=1}^N$ to satisfy two key requirements: (1) $\{\tilde{\pi}_i\}_{i=1}^N$ must support easy sharing of full state information across all agents; and (2) the network must allow efficient training and local action selection by individual agents. Driven by the two requirements, a mixed network architecture involving both global and local components has been developed, as illustrated in Figure \ref{fig-glo-pol}. In this figure, all agents have easy access to vital information embedded in global state through the shared global DNN. Meanwhile, the local network components with respect to individual agents in Figure \ref{fig-glo-pol} simplify the processing required for local action selection. This design also significantly reduces the total number of trainable parameters in the entire network structure. It is interesting to point out that we have adopted clearly a symmetric design to realize both the deterministic part and the stochastic part of $\{\tilde{\pi}_i\}_{i=1}^N$. Despite of the symmetry, the training methods employed for the two parts are different and will be detailed later.

Figure \ref{fig-glo-pol} reverberates the mixing network structure introduced by Q-MIX \cite{rashid2018}. However it can be clearly distinguished from Q-MIX because: (1) our network stands for a model of the global policy rather than the Q-function; and (2) our network processes directly the full state input while Q-MIX incorporates global state information via a separate hyper-network. Despite of the differences, Q-MIX is complimentary to our research and can be utilized as the centralized Q-function in the CTEDD algorithm. However, for the benchmark problems studied in this paper (based on similar benchmarks introduced in \cite{lowe2017nips,mordatch2018}), it is sufficient to model the Q-function via simple feedforward DNNs.

Building on the network design above, we decide to train the policy parameters $\{\theta_i\}_{i=1}^N$ corresponding to the deterministic part in Figure \ref{fig-glo-pol} as well as the policy parameters in the shared global DNN via \eqref{equ-dpg}. For this purpose, we must simultaneously train the centralized Q-network to minimize the approximated loss $\mathcal{L}_{Q}$ defined in \eqref{equ-loss-q}. The learning process closely resembles that of MADDPG. Specifically, for stable learning, CTEDD makes use of the target networks for both the deterministic local DNN and global DNN in Figure \ref{fig-glo-pol} as well as the Q-network.

As for the policy parameters $\{\omega_i\}_{i=1}^N$ associated with the stochastic part in Figure \ref{fig-glo-pol}, they can be trained through \eqref{equ-pgt}. In particular, at every new learning iteration, the most recently collected environment samples since the last learning iteration will form a batch $\mathcal{B}$. Policy gradients can be subsequently estimated based on $\mathcal{B}$\footnote{This is because \eqref{fig-glo-pol} only applies to environment samples obtained by using the policies being trained.}. Although \eqref{equ-pgt} can be used to train both $\{\theta_i\}_{i=1}^N$ and $\{\omega_i\}_{i=1}^N$, to avoid over-training of $\{\theta_i\}_{i=1}^N$, only $\{\omega_i\}_{i=1}^N$ will be trained via \eqref{equ-pgt}. Specifically, aimed at promoting effective and coordinated environment exploration, maximum-entropy DRL techniques will be further employed by CTEDD to train $\{\omega_i\}_{i=1}^N$ along the direction below
\begin{equation}
\begin{split}
&\Delta\omega_i\propto \\
& \frac{1}{\|\mathcal{B}\|}\sum_{\mathcal{B}} \nabla_{\omega_i}\log\tilde{\pi}_i(s_t,\{a^t_j\}_{j=1}^N) Q^{\tilde{\pi}}(s_t,\{a^t_j\}_{j=1}^N) \\
& +\alpha \sum_{\mathcal{B}} \nabla_{\omega_i} \mathcal{H}(\{\tilde{\pi}_j(s_t,\cdot)\}_{j=1}^N)
\end{split},
\label{equ-ome-upt}
\end{equation}
\noindent
where $\alpha$ is the \emph{entropy regularization factor} and $\mathcal{H}(\{\tilde{\pi}_j(s_t,\cdot)\}_{j=1}^N)$ is the Shannon entropy for action sampling across all agents. Moreover $\nabla_{\omega_i}\log\tilde{\pi}_i(s_t,\{a^t_j\}_{j=1}^N)$ can be simplified to $\nabla_{\omega_i}\log\tilde{\pi}_i(s_t,a^t_i)$ and $\nabla_{\omega_i} \mathcal{H}(\{\tilde{\pi}_j(s_t,\cdot)\}_{j=1}^N)$ can be simplified to $\nabla_{\omega_i} \mathcal{H}(\tilde{\pi}_i(s_t,\cdot))$, since $\omega_i$ only affects local action selection by agent $A_i$. Meanwhile, we replace $Q^{\tilde{\pi}}(s_t,\{a^t_j\}_{j=1}^N)$ in \eqref{equ-ome-upt} by
\[
Q^{\tilde{\pi}}(s_t,\{a^t_j\}_{j=1}^N)-Q^{\tilde{\pi}}(s_t,\{\pi_{\theta_j}(s_t)\}_{j=1}^N).
\]
\noindent
Here $\{a^t_j\}_{j=1}^N$ refers to the sampled actions and $\{\pi_{\theta_j}(s_t)\}_{j=1}^N$ denotes the actions identified by the deterministic parts of the global policies. Apparently, if the subtraction above gives positive outcomes, exploring more of the environment is considered preferable. Conversely, if the subtraction produces negative values, random action sampling will be discouraged. Through this way, agents can collaboratively control their environment exploration based on the current state as well as past experiences.
\begin{figure}[ht!]
\centering
\includegraphics[width=0.8\columnwidth]{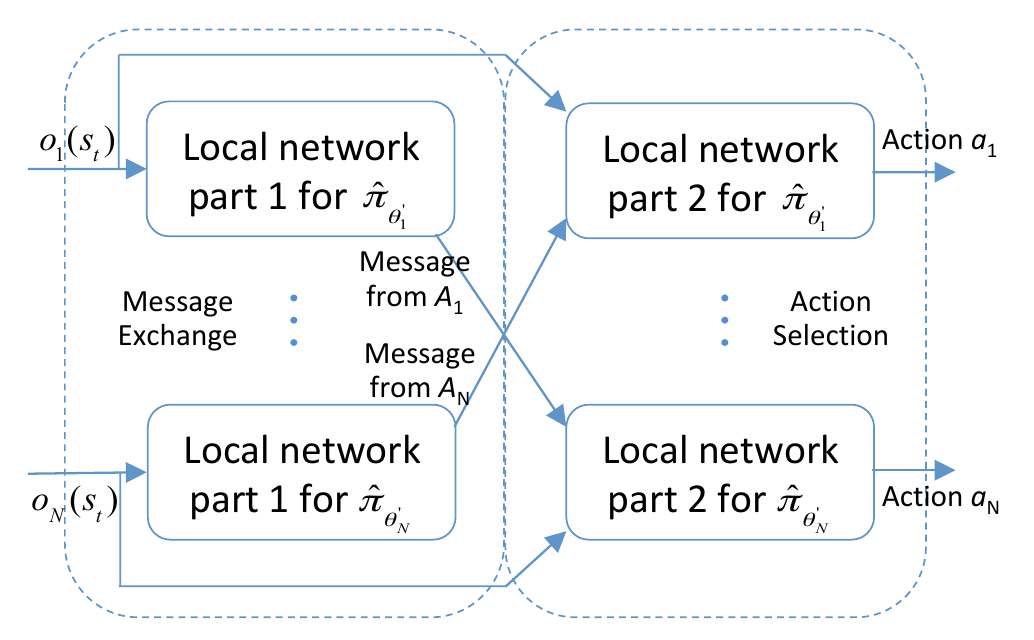}
\caption{The local policy network design with support for inter-agent communication.}
\label{fig-loc-pol}
\end{figure}

\subsection{Learning Local Policies via Policy Distillation}
\label{sub-met-pd}

To fulfill the learning goal in \eqref{eq-pro-def}, the second stage of CTEDD builds locally executable policies for every agent from the trained global policies. Our algorithm design can work straightforwardly with arbitrary network architectures for the local policies and support various inter-agent communication methods. However without exploring complicated communication technologies, similar to many previous works \cite{foerster2016comm,sukhbaatar2016}, simple agent-to-agent communication channels will be utilized in our experiments to allow each agent to send a small vector of information to all other agents at every time step. The network design of the local policies and the communication channels are depicted in Figure \ref{fig-loc-pol}.

As shown in Figure \ref{fig-loc-pol}, the local policy networks $\{\hat{\pi}_{\theta'_i}\}_{i=1}^N$ consist of two successive parts. Part 1 is responsible for message generation based on agents' local observations. The generated messages are subsequently transferred to all other agents. Following message exchange, every agent $A_i$ activates part 2 of its policy network to produce the final action output $a_i$ based on both its local observation $o_i(s_t)$ as well as all messages received. The whole process happens within every time step and must be very efficient. For this purpose, we only allow small messages to be exchanged among agents, making it difficult to directly learn highly coordinated local policies without any guidance from global policies.

As described in Algorithm \ref{alg-1}, local policies will be trained based on state samples collected previously by the global policies. Specifically we order all state samples chronologically according to when they were sampled. State samples obtained at earlier times will be used first to train local policies, followed by samples collected at later times. Chronological sampling for policy distillation is important to ensure that the sampling distribution is not too different from the distilled local policies in order for them to better approximate their global counterparts\footnote{We found experimentally that this chronological approach enables local policies to achieve better performance in comparison to randomly sampling state samples from the whole replay buffer.}. For every batch of samples $\mathcal{B}$ just retrieved from the buffer, the policy parameters $\{\theta'_i\}_{i=1}^N$ of all local policies will be trained for a few iterations to minimize the loss $\mathcal{L}_D$ defined in \eqref{equ-loss-d} below
\begin{equation}
\mathcal{L}_D=\frac{1}{\|\mathcal{B}\|}\sum_{\mathcal{B}} \sum_{i=1}^N \left\|\pi_{\theta_i}(s_t)-\hat{\pi}_{\theta'_i}(o_i(s_t),\{m_{j\neq i}\}_{j=1}^N)\right\|_2^2.
\label{equ-loss-d}
\end{equation}
\noindent
Afterwards, a new batch will be retrieved and utilized to drive the next learning iteration. This process will continue until either all samples in the buffer have been consumed or the testing performance of the trained local policies is considered satisfactory (or cannot be improved further). For the former case which is adopted in our experiments, no extra environment sampling is required for policy distillation and the sample efficiency of CTEDD can hence be guaranteed.

\section{Experiments}
\label{sec-exp}

We introduce four environments (i.e., MADRL benchmarks) to examine empirically the sample efficiency and performance of CTEDD, with MADDPG serving as the baseline algorithm in all our experiments\footnote{We did not experimentally compare CTEDD with single-agent learning algorithms such as DDPG and TRPO or MADRL algorithms such as COMA because they have already been shown to be less effective than MADDPG \cite{lowe2017nips,iqbal2018}. We have experimented several variations of MADDPG including MADDPG+SAC utilized in \cite{iqbal2018}. However MADDPG+SAC does not perform better than MADDPG.}. Detailed introduction to each environment can be found in Subsection \ref{sub-sec-prob}. All competing algorithms have been evaluated in regard to two different settings, namely, agents can exchange messages in the form of either one-dimensional or three-dimensional continuous vectors with each other at every time step. More discussion on the hyper-parameter settings of these algorithms can be found in Subsection \ref{sub-par-set}.

\subsection{Benchmark Environments}
\label{sub-sec-prob}

\begin{figure}[ht!]
\centering
\includegraphics[width=0.45\columnwidth]{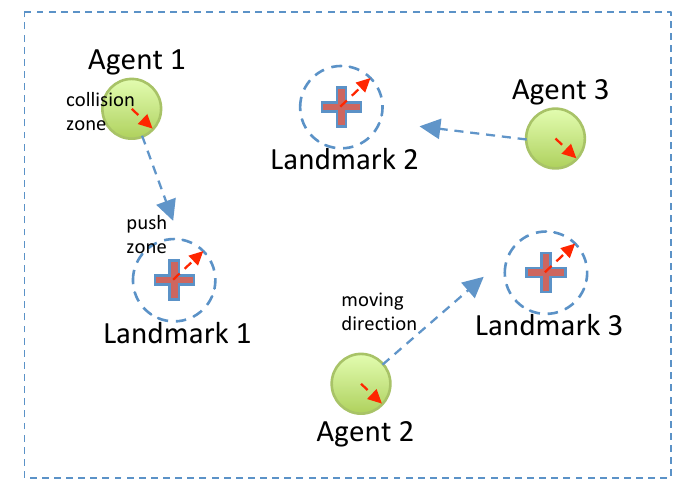}
\caption{An illustration of the experimental environments, involving multiple agents and landmarks. Each agent has a collision zone and can move freely in any direction in the two-dimensional world. Each landmark is associated with a push zone in the Push Forward environment.}
\label{fig-env-exp}
\end{figure}

All the four benchmark environments studied in this paper adopt the \emph{multi-agent particle environment framework} proposed and extensively utilized in \cite{mordatch2018,lowe2017nips}. This framework is suitable for building continuous MADRL problems that demand for complex inter-agent coordination while allowing simple control of agents' behaviors and their local perception. In all the four environments, there are multiple agents living jointly in a two-dimensional continuous world, as illustrated in Figure \ref{fig-env-exp}. An individual agent cannot solve the given problems single handedly and requires collaborative efforts from its teammates. In addition, multiple (two or three) landmarks will be deployed as problem solving targets. Unlike agents, landmarks either remain still or can update their locations from time to time based on some pre-defined rules.

\begin{figure*}[!ht]
\center
      \begin{minipage}[t]{0.246\textwidth}
        \includegraphics[width=\textwidth]{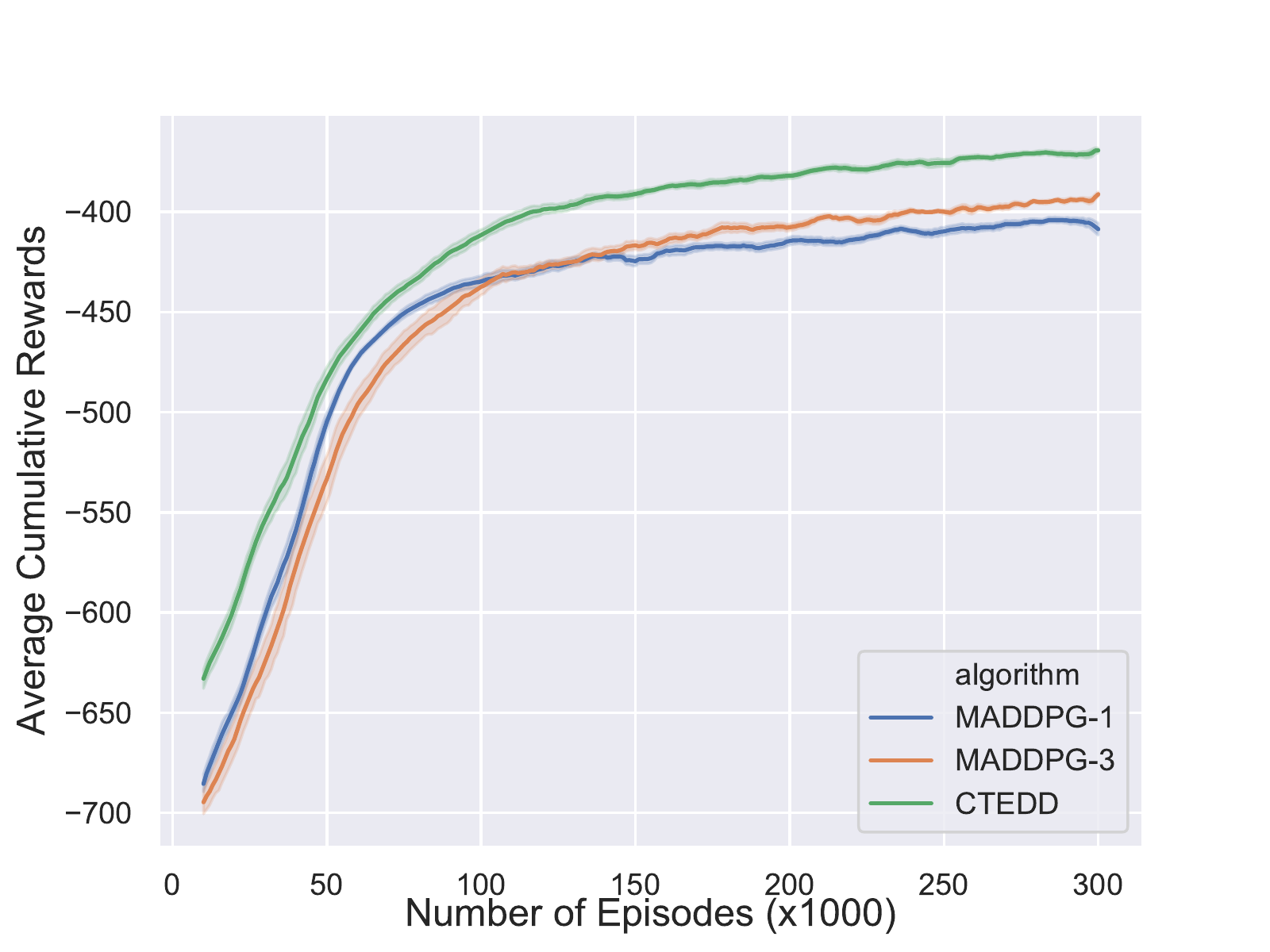}
        \subcaption{CN-V1}
      \end{minipage}%
      \begin{minipage}[t]{0.246\textwidth}
        \includegraphics[width=\textwidth]{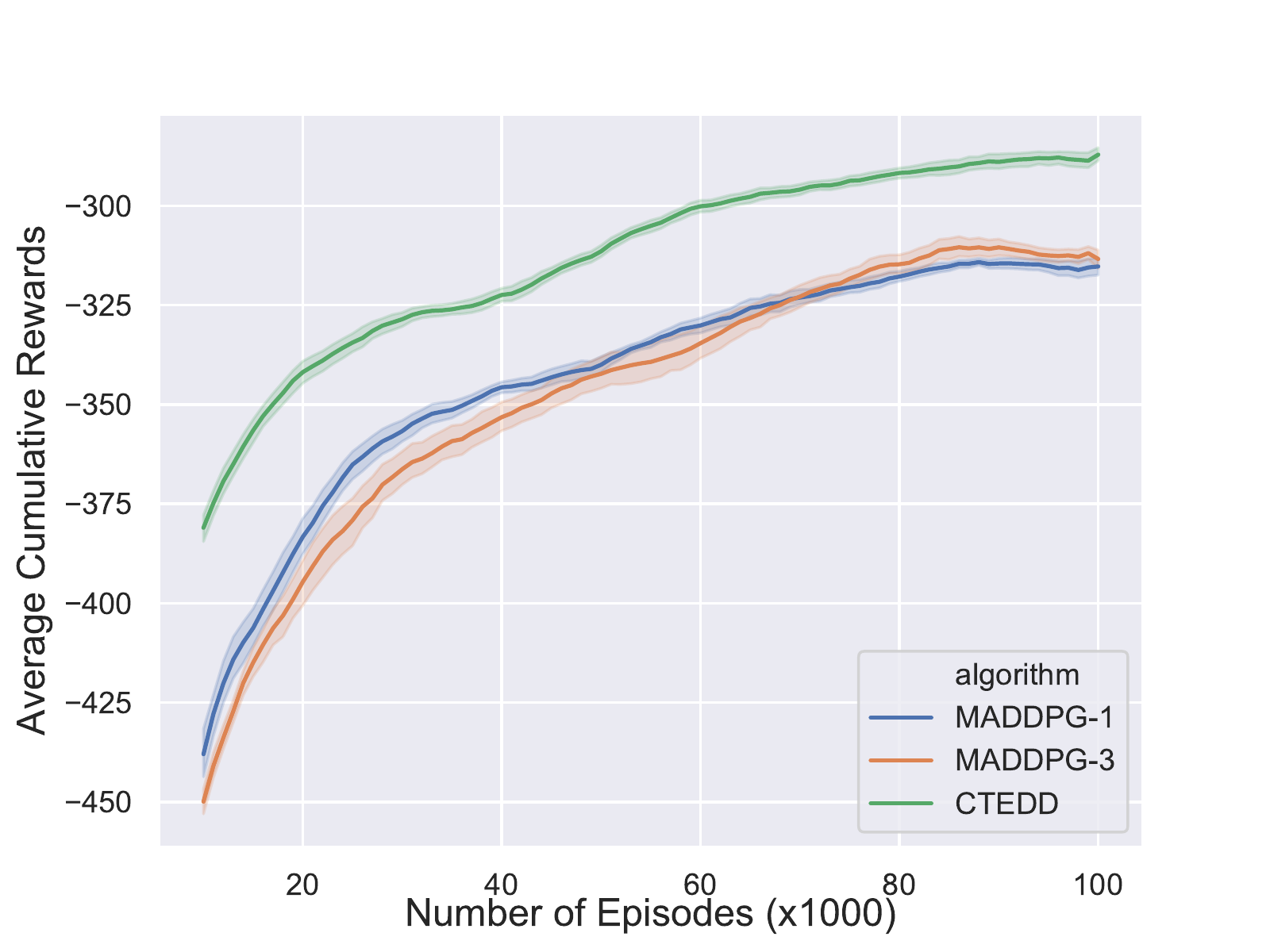}
        \subcaption{CN-V2}
      \end{minipage}
      \begin{minipage}[t]{0.246\textwidth}
        \includegraphics[width=\textwidth]{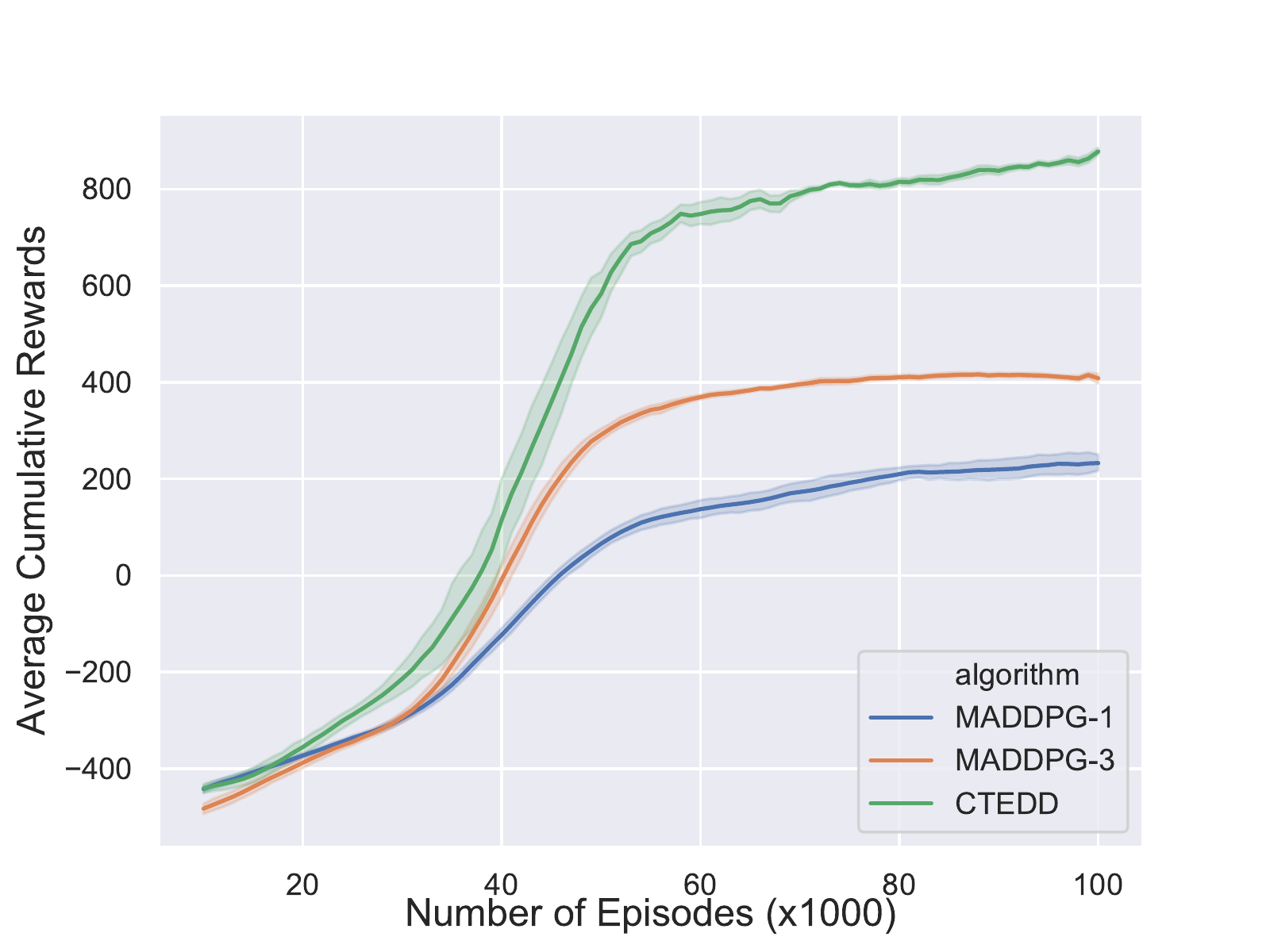}
        \subcaption{PF}
      \end{minipage}
      \begin{minipage}[t]{0.246\textwidth}
        \includegraphics[width=\textwidth]{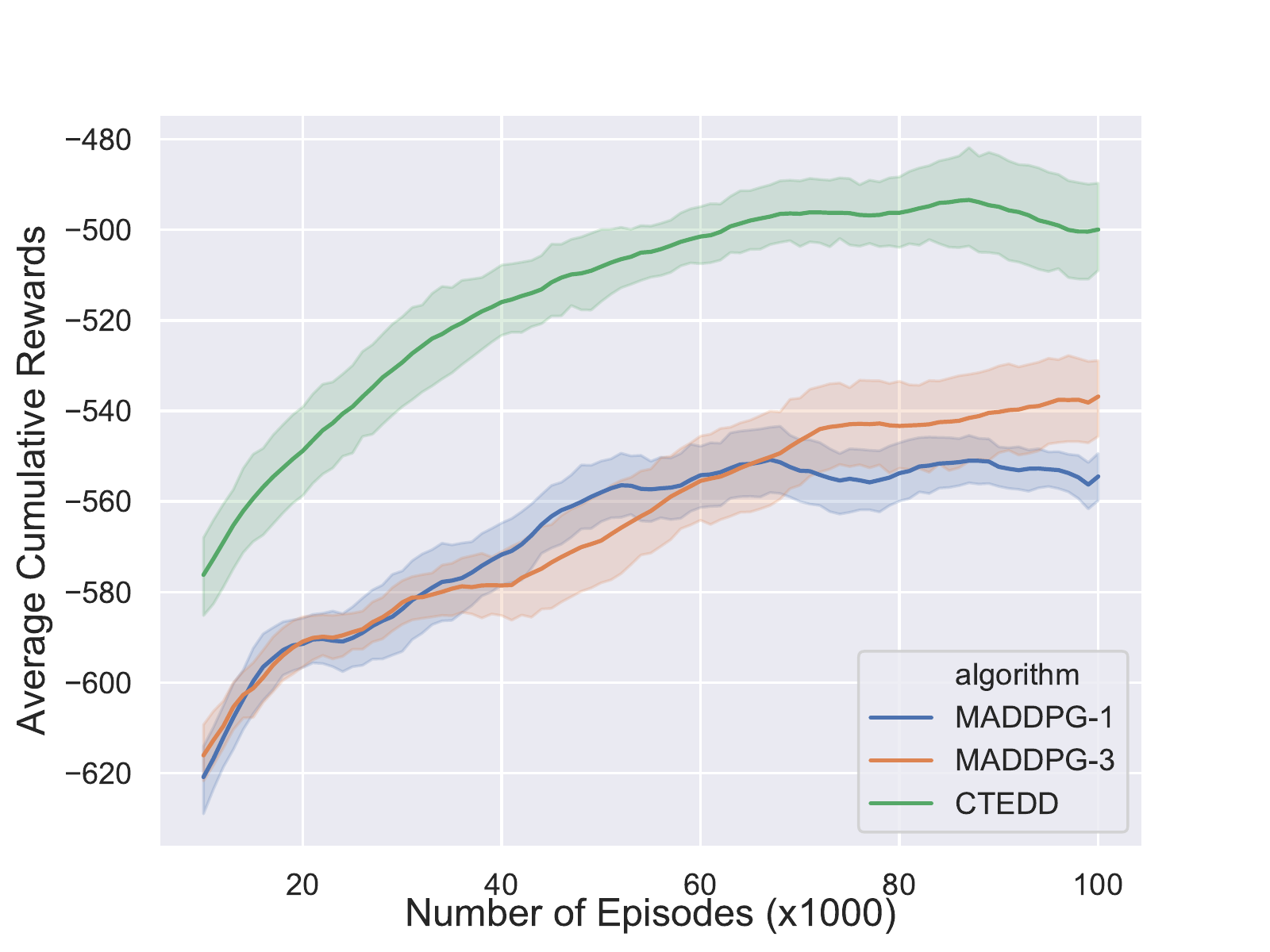}
        \subcaption{PP}
      \end{minipage}
\caption{The learning performance of CTEDD, MADDPG-1 and MADDPG-3 on four environments, namely CN-V1, CN-V2, PF and PP.}
\label{fig-res-1}
\end{figure*}

\begin{figure*}[!ht]
\center
      \begin{minipage}[t]{0.246\textwidth}
        \includegraphics[width=\textwidth]{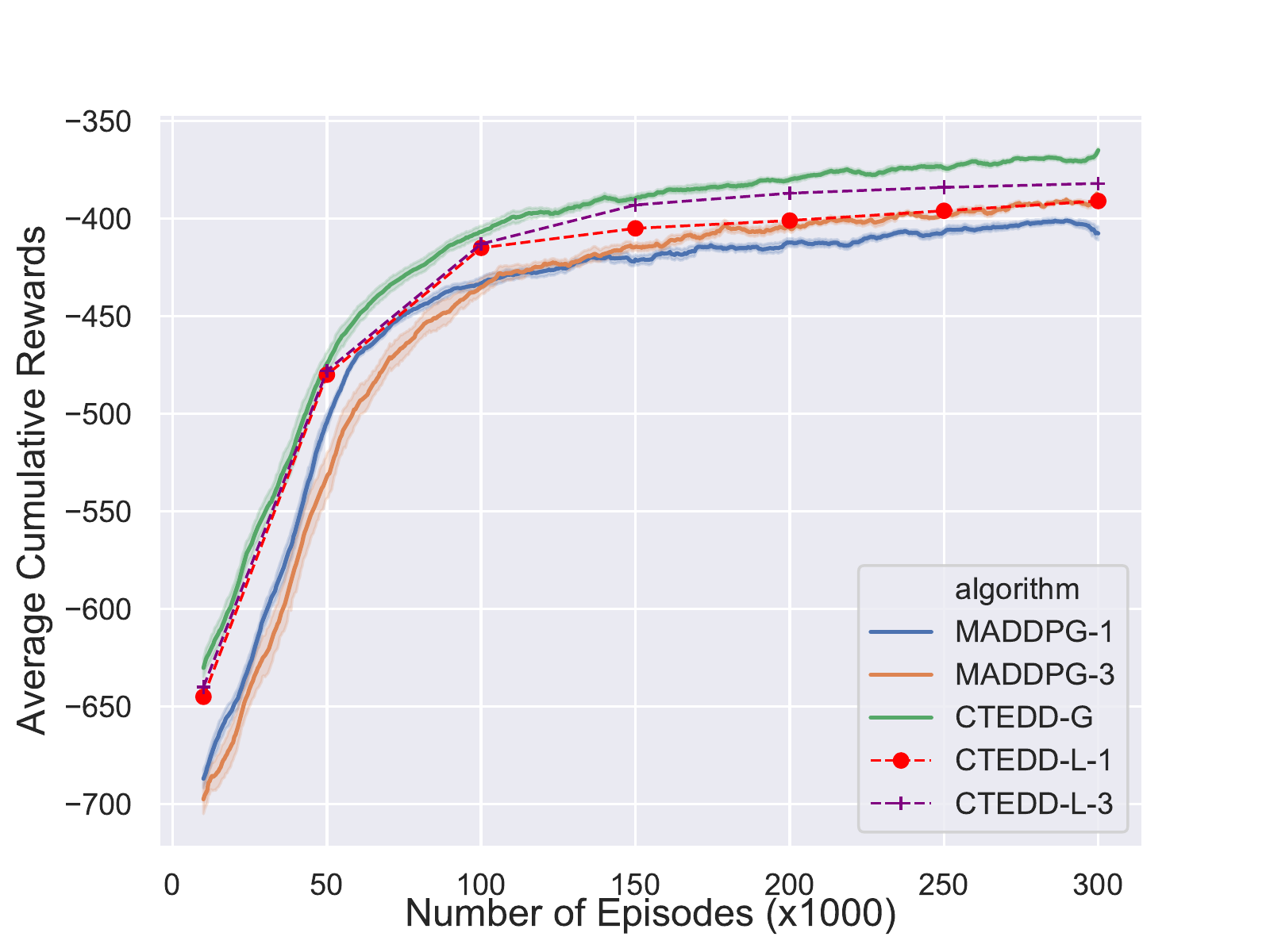}
        \subcaption{CN-V1}
      \end{minipage}%
      \begin{minipage}[t]{0.246\textwidth}
        \includegraphics[width=\textwidth]{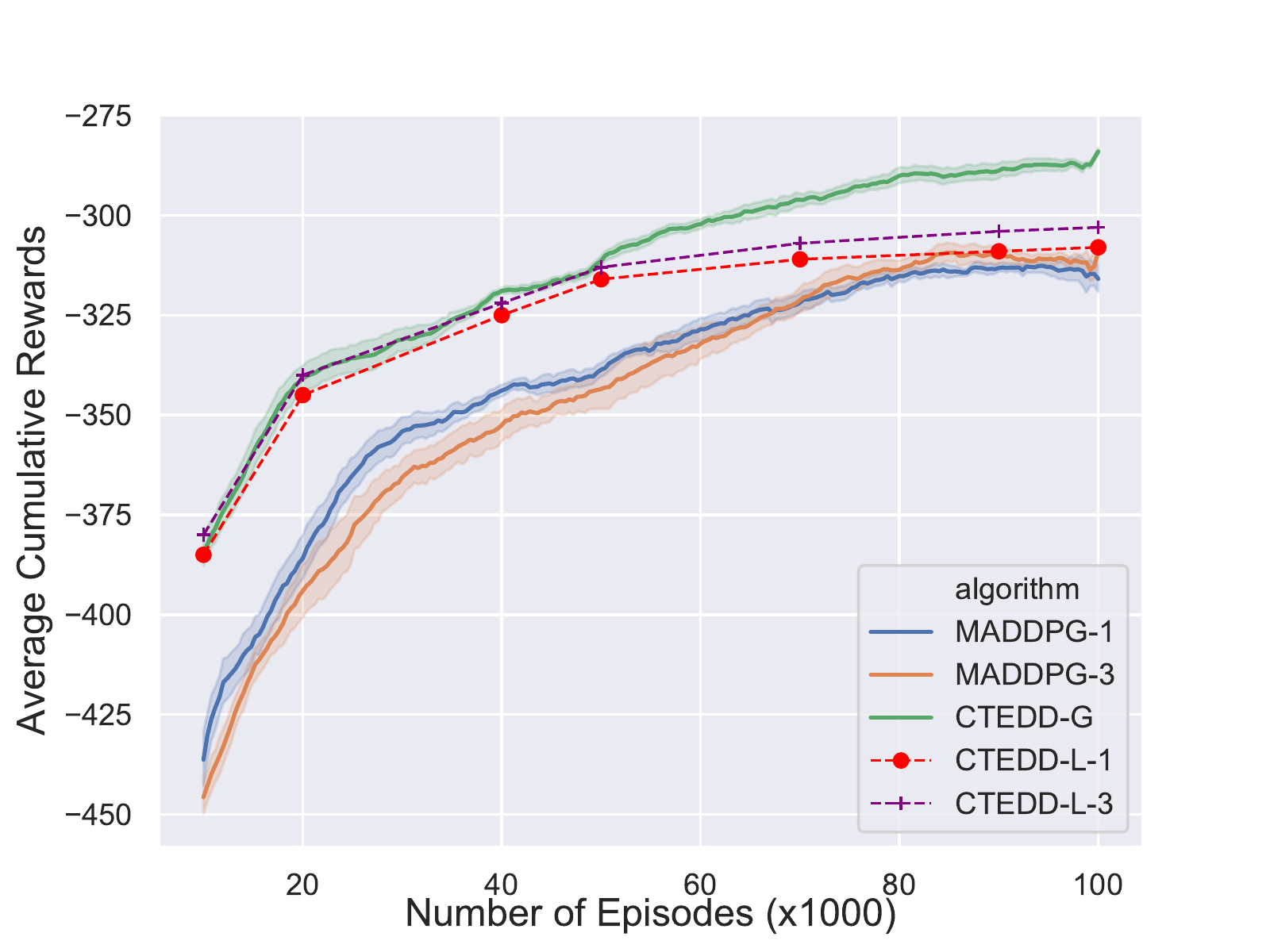}
        \subcaption{CN-V2}
      \end{minipage}
      \begin{minipage}[t]{0.246\textwidth}
        \includegraphics[width=\textwidth]{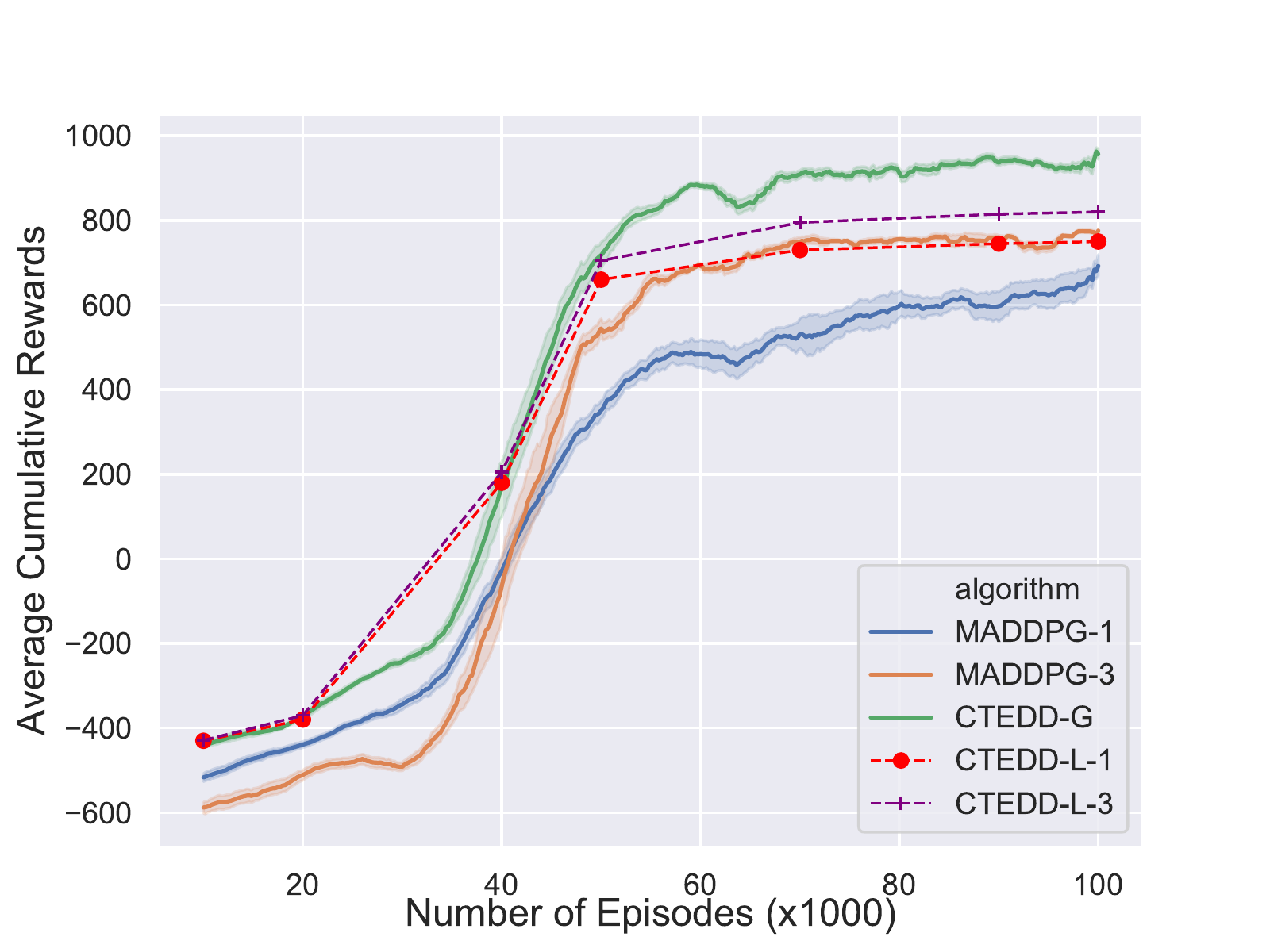}
        \subcaption{PF}
      \end{minipage}
      \begin{minipage}[t]{0.246\textwidth}
        \includegraphics[width=\textwidth]{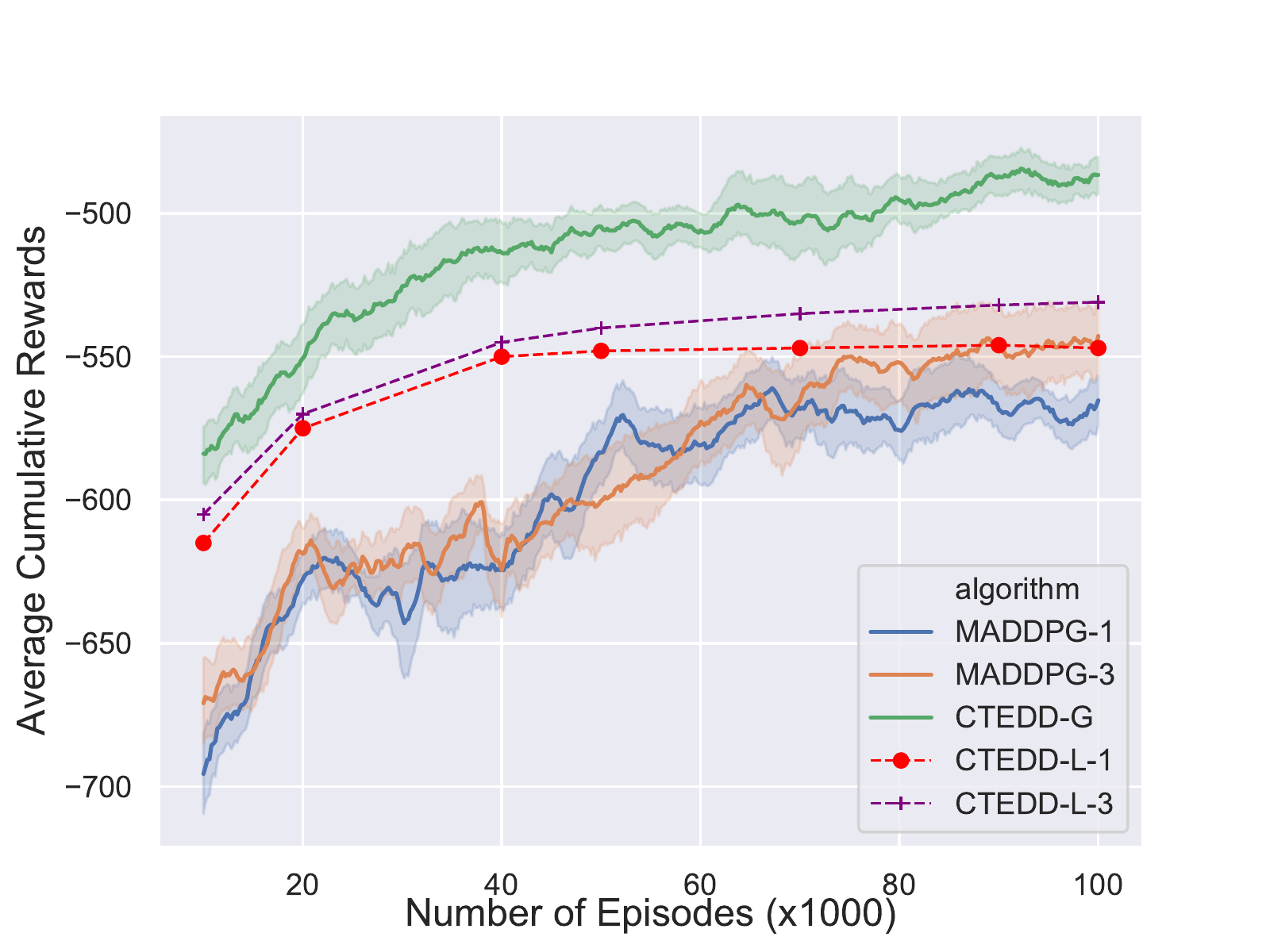}
        \subcaption{PP}
      \end{minipage}
\caption{The testing performance of CTEDD-G, CTEDD-L-1, CTEDD-L-3, MADDPG-1 and MADDPG-3 on four environments, namely CN-V1, CN-V2, PF and PP.}
\label{fig-res-2}
\end{figure*}

\subsubsection*{Cooperative Navigation V1}

This environment, known as {\bf CN-V1}, is constructed based on a similar environment introduced in \cite{lowe2017nips}. There are three agents and three static but randomly located landmarks in this environment. Every agent must coordinate with each other to occupy a separate landmark. When more than one agents occupy the same landmark simultaneously, a collision will occur because every agent is at the centre of a separate collision zone associated with significant physical space, causing a penalty of -1.0 for all agents involved in the collision. Additionally, the step-wise reward also depends on the shortest distance between every landmark and one of the agent in the environment. To achieve the learning goal and maximize the cumulative rewards, at every time step, we allow any individual agent to measure its Euclidean distance from other agents and all the landmarks. Such local observation does not tell the velocity and moving direction of other agents. It does not tell the actual location of any landmark either and hence only serves as a partial view of the multi-agent world. Based on the local observation and short messages (i.e., low-dimensional vectors) exchanged with other agents, the three agents must learn to infer the landmarks to be occupied by each and try to avoid potential collisions. Every problem episode lasts for 25 consecutive time steps.

\subsubsection*{Cooperative Navigation V2}

This environment, known as {\bf CN-V2}, is a variation of CN-V1. Instead of having three landmarks, CN-V2 has two landmarks deployed at random locations. Again, three agents are targeted to occupy both of the two landmarks subject to some extra restrictions. First, if landmark 1 has shorter distance to any agent than landmark 2, then two agents must be closer in distance to landmark 1 and one agent must be more closer to landmark 2. If two agents are actually located closer to landmark 2, a penalty of -10.0 will be incurred. The case when landmark 2 has shorter distance to any agent is treated in the same manner. Identical to CN-V1, any collision will cause a penalty of -1.0 for all collided agents. Due to this reason, while two agents are trying to occupy the same landmark, they must manage to stay in a safe distance away from each other to prevent collision. We provide the same local observations for all agents, as implemented in CN-V1. Every problem episode lasts for 25 consecutive time steps.

\subsubsection*{Push Forward}

This environment, known as {\bf PF}, features three collaborating agents and three movable landmarks. The learning goal is for all agents to jointly push the three landmarks horizontally to the right (along the $X$ axis). For this purpose, each of the three agents must first move into the push zone of a different landmark. When all agents are in the respective push zones at the same time, they must simultaneously move towards the right direction (i.e., the location of these agents at the next time step must be to the right of their previously location). If so, the three landmarks will be pushed to the right by a small distance and a reward of +10.0 will be offered to every pushing agent. Meanwhile, the reward will be deducted by the distance of every landmark from its closest agent. The collision penalty also applies whenever two or more agents collide on their ways. Every agent adopts the same local observation as usual (see CN-V1). A problem episode lasts for 50 consecutive time steps.

\subsubsection*{Predator and Prey}

This environment, known as {\bf PP}, contains three chasing agents (i.e., predators) and two movable landmarks (i.e., preys). The goal is for agents to catch one or both landmarks. For this to happen, all the three agents must move concurrently to the close proximity of one or two landmarks. In other words, a landmark can only be captured by three agents together. Upon capturing a landmark, a reward of +10.0 will be provided by the learning environment. Meanwhile, while chasing a landmark, the agents must keep a safe distance from each other to prevent collision. Similar to PF, the reward will also be deducted by the distance between every landmark and its closet agent. To escape from a capture, we build in man-made rules in each landmark to update its location at every time step. Specifically, we assume that a landmark has global knowledge of the multi-agent world. Based on each agent's current location, velocity and acceleration, the landmark can predict the future locations of all agents. Accordingly, the landmark will choose to move by a fixed distance along the direction that will maximize the total distance between the landmark and all the agents. Apparently, landmarks are much more agile than agents since their next moves are not influenced by their current velocity and acceleration, making them difficult to catch if agents fail to collaborate effectively. The same local observation as described above is accessible to each agent. A problem episode lasts for 50 consecutive time steps.

\begin{figure*}[!ht]
\center
      \begin{minipage}[t]{0.25\textwidth}
        \includegraphics[width=\textwidth]{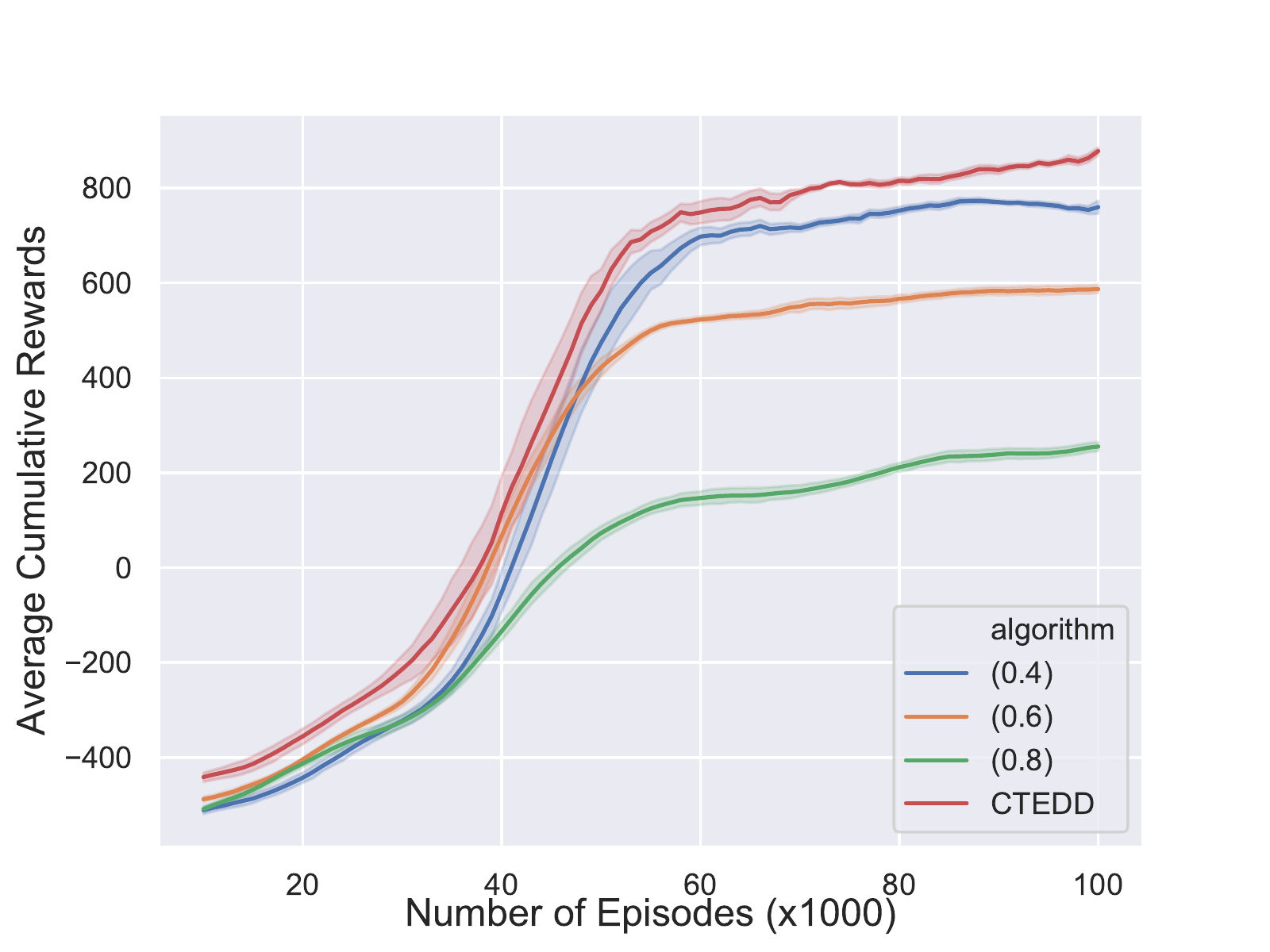}
        \subcaption{Learning Performance}
      \end{minipage}%
      \begin{minipage}[t]{0.25\textwidth}
        \includegraphics[width=\textwidth]{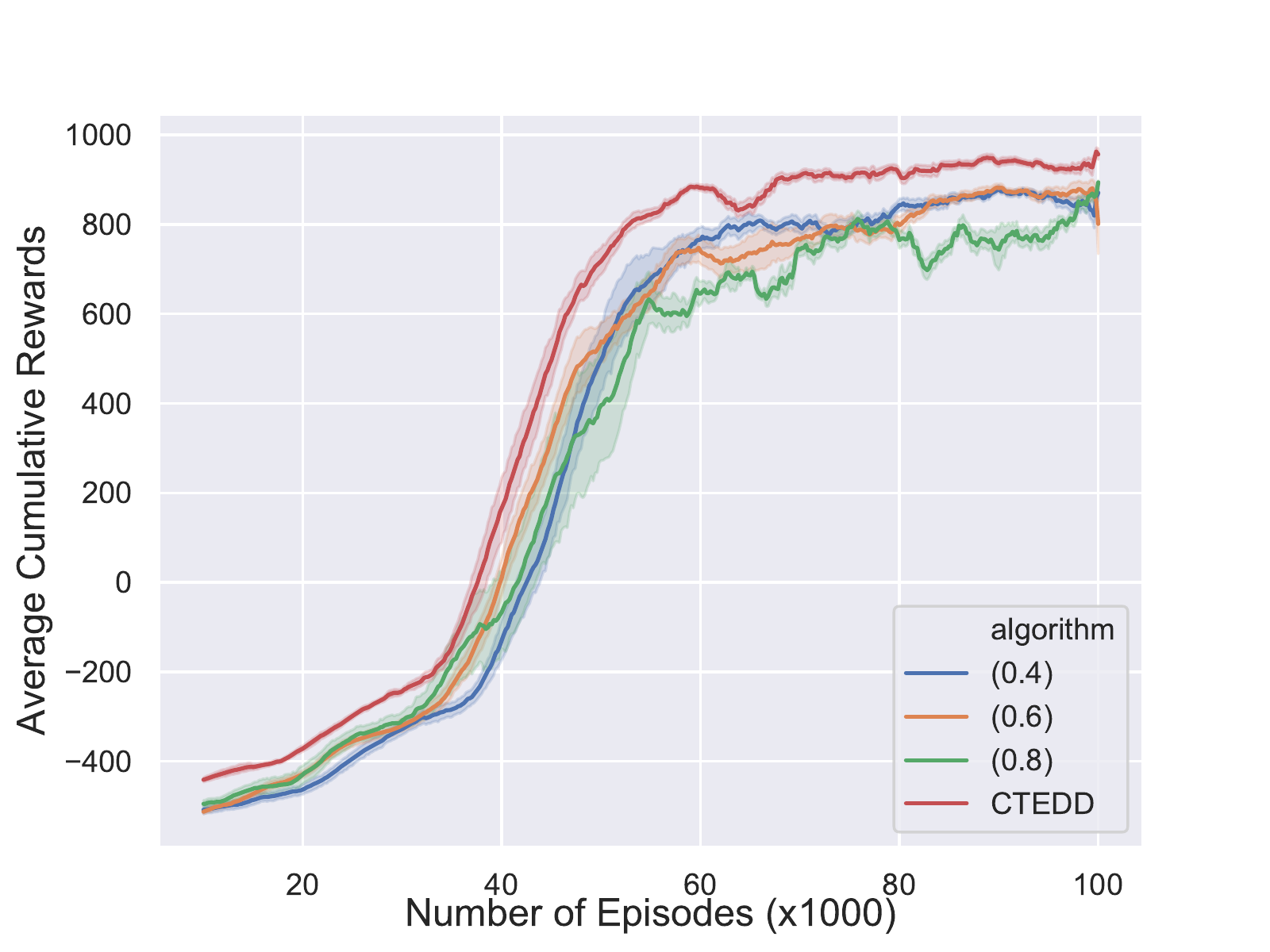}
        \subcaption{Testing Performance}
      \end{minipage}
\caption{The learning and testing performance of CTEDD on PF when agents sample their actions with fixed level of standard deviations during training. Three different levels have been considered. They are indicated respectively as (0.4), (0.6) and (0.8) in this figure. CTEDD stands for the case when the standard deviations for action sampling are further trained by using \eqref{equ-ome-upt}.}
\label{fig-res-3}
\end{figure*}

\subsection{Algorithm Hyper-Parameter Settings}
\label{sub-par-set}

In our experiments, the global feedforward network in Figure \ref{fig-glo-pol} contains one hidden layer with 64 hidden units. Each local network component in the same figure contains 64 hidden units located in a single hidden layer too. As for the local policies trained directly by MADDPG and depicted in Figure \ref{fig-loc-pol}, both part 1 and part 2 of the corresponding policy networks contain a single hidden layer made up of 64 units. Such local policies will be trained by CTEDD as well during the policy distillation stage. Different from policy networks, the Q-function is consistently approximated by DNNs with $64\times 64$ hidden units in all competing algorithms.

The learning rate for MADDPG is set to 0.01 which is the default setting in the source code provided by MADDPG inventors\footnote{https://github.com/openai/maddpg}. As for CTEDD, we tested several learning rates including 0.02, 0.01, 0.005 and 0.002 but did not notice significant impacts on learning performance. However CTEDD appears to be slightly more reliable when the learning rate is set to 0.005. Similar tests have been conducted on MADDPG too, confirming that 0.01 is a good choice for the learning rate. Besides, we adopt an annealing schedule for the entropy regularization factor $\alpha$ in \eqref{equ-ome-upt}. Specifically, $\alpha$ is set initially to 0.1 and gradually decremented to 0.001 over 3M consecutive sampling steps. Afterwards, it will remain at 0.001 till the end of the learning process. We have also tried to fix $\alpha$ at 0.1, 0.01 and 0.001 respectively while training global policies and discovered that the annealing schedule can induce a better balance between exploration and exploitation. Other than the above, all other hyper-parameters are made identical for both CTEDD and MADDPG, according to the default settings in the MADDPG source code. Particularly, the batch size is 1024 and $\gamma=0.95$.

\subsection{Experiment Results}

Figure \ref{fig-res-1} compares the learning performance of CTEDD, MADDPG-1 and MADDPG-3 on all the four environments. To ensure that the observed performance differences are statistically significant, every algorithm has been tested on each environment with 10 different random seeds. Both the average learning performance (measured by the average cumulative rewards obtained jointly by all agents in a problem episode) and its confidence interval have been depicted in Figure \ref{fig-res-1}. We can draw two major conclusions from this figure. First, CTEDD can clearly learn faster and achieve higher effectiveness than MADDPG. In fact, CTEDD consistently outperformed MADDPG-1 and MADDPG-3 on all environments, confirming that the global policies trained by CTEDD can promote coordinated environment exploration and learning among all agents. Second, exchanging more information among the learning agents can contribute positively to the final learning performance. But this may slow down the learning process. For example, on CN-V1 and CN-V2, MADDPG-3 improved slowly at the beginning but eventually surpassed MADDPG-1 in learning performance. This is easily understandable since MADDPG-3 requires agents to communicate more information with each other, inevitably increasing the complexity of the local policy networks. However, this will enable agents to learn to cooperate more effectively at a later stage.

Besides the learning performance, Figure \ref{fig-res-2} further compares the testing performance of CTEDD, MADDPG-1 and MADDPG-3. In our experiments, after obtaining every 10,000 environment samples, the policies learned so far will be tested in separate testing environments which have been created as identical copies of the training environments. A total of 400 testing episodes will be carried out and the average cumulative rewards obtained across these episodes represent the testing performance of the trained policies. In regard to CTEDD, we have tested both the trained global policies (only the deterministic parts) as well as the local policies obtained through policy distillation. These two cases are indicated separately as CTEDD-G and CTEDD-L in Figure \ref{fig-res-2}. CTEDD-L also has two variations, i.e., CTEDD-L-1 and CTEDD-L-3, depending respectively on whether distilled local policies allow agents to exchange one-dimensional or three-dimensional messages.

As clearly shown in Figure \ref{fig-res-2}, on all environments, the performance of CTEDD-L-1 and CTEDD-L-3 follows closely with that of CTEDD-G up to a certain level. Afterwards, CTEDD-L-1 and CTEDD-L-3 can no longer noticeably improve their performance any further. This observation agrees well with our expectation since the level of inter-agent coordination (hence the testing performance) depends heavily on agents' ability to access shared global information. In comparison to MADDPG-1 and MADDPG-3, we can also conclude that CTEDD-L-1 and CTEDD-L-3 require significantly less number of learning episodes to reach the same level of performance. For example, on CN-V1, CTEDD-L-3 can reach the average testing performance of -390 after approx. 150,000 learning episodes. On the other hand, MADDPG-3 requires twice the number of episodes. Meanwhile, as demonstrated in Figure \ref{fig-res-2}, the same global policy learned by CTEDD can be employed to train different local policies supported by varied communication channels (e.g. one-dimensional or three-dimensional channels), without consuming extra environment samples. CTEDD is hence more sample efficient and flexible than MADDPG\footnote{While we focus on sample efficiency, global policy training and local policy distillation can actually be performed in parallel in CTEDD. Hence, with parallel computing facilities, the time efficiency is not an issue for CTEDD.}.

We have also investigated the effectiveness of the maximum-entropy RL technique adopted in \eqref{equ-ome-upt}. Using the PF environment as an example, Figure \ref{fig-res-3} reports both the learning and testing performance of CTEDD when agents sample their actions with fixed level of randomness during training. As clearly evidenced in this figure, without \eqref{equ-ome-upt}, it is hard to set the standard deviation for action sampling properly. Setting it to a high value (e.g., 0.8) will hurt the final performance of CTEDD. On the other hand, when the standard deviation is set to a low level (e.g., 0.4), the initial learning speed may suffer (in comparison to the case when the standard deviation equals to 0.6 or 0.8). In contrast, by utilizing \eqref{equ-ome-upt}, CTEDD can manage to achieve the best learning speed and performance. Meanwhile, a comparison to experiment results of MADDPG depicted in Figures \ref{fig-res-1} and \ref{fig-res-2} indicate that CTEDD can still outperform MADDPG without using any maximum-entropy RL methods.

\section{Conclusions}
\label{sec-con}

Our research in this paper was inspired by the fact that effective DRL in complex multi-agent systems demand for highly coordinated environment exploration among all learning agents. This notion drove us to propose a new CTEDD framework to promote easy and effective sharing of global information. Our idea was realized specifically by applying DDPG to training global policies approximated as DNNs with mixed local and global components. We have also utilized the maximum-entropy learning technique to properly balance the trade-off between exploration and exploitation, further promoting coordinated action sampling and environment exploration across all agents. Meanwhile, a policy distillation technique was adopted by us to derive locally executable policies for each agent from well-trained global policies in a highly sample efficient manner. As evidenced in our experiments, the newly proposed CTEDD framework and the corresponding algorithm can significantly improve the learning performance and sample efficiency on several MADRL benchmarks, in comparison to the state-of-the-art MADDPG algorithm.

\bibliographystyle{ACM-Reference-Format}
\bibliography{citefile}

\end{document}